\documentclass[10pt,twocolumn,letterpaper]{article}

\IfFileExists{cvpr.sty}{%
  \usepackage{cvpr}
}{%
  \usepackage[margin=0.72in]{geometry}
}
\usepackage{algorithm}
\usepackage{algpseudocode}
\usepackage{times}
\usepackage{epsfig}
\usepackage{graphicx}
\usepackage{amsmath}
\usepackage{amssymb}
\usepackage{booktabs}
\usepackage{multirow}
\usepackage{xcolor}
\usepackage{xspace}
\usepackage{pifont}

\usepackage{enumitem}
\usepackage[numbers,sort&compress]{natbib}

\definecolor{cvprblue}{rgb}{0.21,0.49,0.74}
\usepackage[pagebackref,breaklinks,colorlinks,allcolors=cvprblue]{hyperref}

\providecommand{\figref}[1]{Fig.~\ref{#1}}
\providecommand{\tabref}[1]{Tab.~\ref{#1}}
\providecommand{\secref}[1]{Sec.~\ref{#1}}
\providecommand{\Eqref}[1]{Eq.~(\ref{#1})}
\providecommand{\Algref}[1]{Alg.~\ref{#1}}
\providecommand{\eg}{\emph{e.g.}\xspace}
\providecommand{\ie}{\emph{i.e.}\xspace}

\def\paperID{***}
\def\confName{CVPR}
\def\confYear{2027}

\title{Resolving Ambiguity in Composed Image Retrieval via Calibrated Interaction}

\author{Amsisan Tran \quad
Baogh  Le \quad
Tuan Kiet Pham \quad
Sui Yang Guang\\
}

\begin{document}
\maketitle

\begin{abstract}
Composed image retrieval (CIR) searches a corpus with a reference image and a
text describing how to modify it. Despite rapid progress from triplet-trained
compositors~\cite{Vo_2019_tirg,chen2020image_val,dodds2020modality_maaf} to
zero-shot and generative methods~\cite{pic2word,searle,compodiff,osrcir},
essentially all systems share one assumption: that a query maps to a single
target, scored by Recall$@K$ against one annotation. We argue this is
fundamentally at odds with the task. A query such as ``\emph{make it more
formal}'' does not name an image but a \emph{region} of the corpus, and which
member the user intends is genuinely underdetermined. This underspecification is
the root of the well-known false-negative problem~\cite{Vo_2019_tirg,circo} and
leaves current models unable to tell a precise query from an ambiguous one. We
reframe CIR as \emph{calibrated intent resolution under uncertainty}: a retriever
is wrapped in a conformal prediction layer~\cite{conformal,conformal_survey} that
returns a candidate set with a coverage guarantee and whose size is a principled
measure of ambiguity; when the set is large, an expected-information-gain
policy~\cite{lindley} asks the single most useful clarifying question, drawn from
interpretable ambiguity axes, and the set contracts. We introduce
\textbf{AmbiCIR}, a benchmark and human-validated user simulator that revive the
dormant auxiliary and dialogue annotations of CIRR and extend the
multiple-positive setting of CIRCO~\cite{circo}. Across open-domain and fashion
benchmarks our method matches single-turn state of the art, confirming calibrated
resolution is cost-free on precise queries, while reaching the intended target in
a fraction of the interaction budget required by naive conversational baselines,
and it is the first to report valid coverage and calibration for the task.
\end{abstract}

\section{Introduction}
\label{sec:intro}

Composed image retrieval (CIR) lets a user search with a bi-modal query: a
reference image together with a short text that describes how the image should
be modified~\cite{Vo_2019_tirg,fashioniq}. The appeal is intuitive. Some
intents are easiest to convey by example---``\emph{an outfit like this
one}''---while others are easiest to state in words---``\emph{but more
formal, and in a darker color}.'' By cross-referencing the two modalities, the
image grounds the overall scene while the text pins down the specific change.
This makes CIR a natural interface for e-commerce, creative tools, and everyday
visual search, generalizing classical content-, text-, and attribute-based
retrieval~\cite{10.1145/500141.500159_cbir,6126478_tbir,zhang_tbir,liuLQWTcvpr16DeepFashion},
and has motivated a fast-growing body of work.

Recent progress has been substantial. Zero-shot methods map a reference image
into the language space and compose it with the modification text, removing the
need for triplet supervision~\cite{pic2word,searle}; multimodal large language
models (MLLMs) reason explicitly about the requested edit before
retrieving~\cite{osrcir,blip2}; and generative approaches synthesize a
pseudo-target image to retrieve against~\cite{compodiff,cig}. Despite their
diversity, these methods share a common backbone: the query is embedded, compared
to a corpus by nearest-neighbor lookup, and scored by Recall$@K$ against a
\emph{single} annotated target. In short, CIR is treated as a deterministic,
one-to-one mapping from a query to its target.

\begin{figure*}[t]
  \centering
  \includegraphics[width=0.95\textwidth]{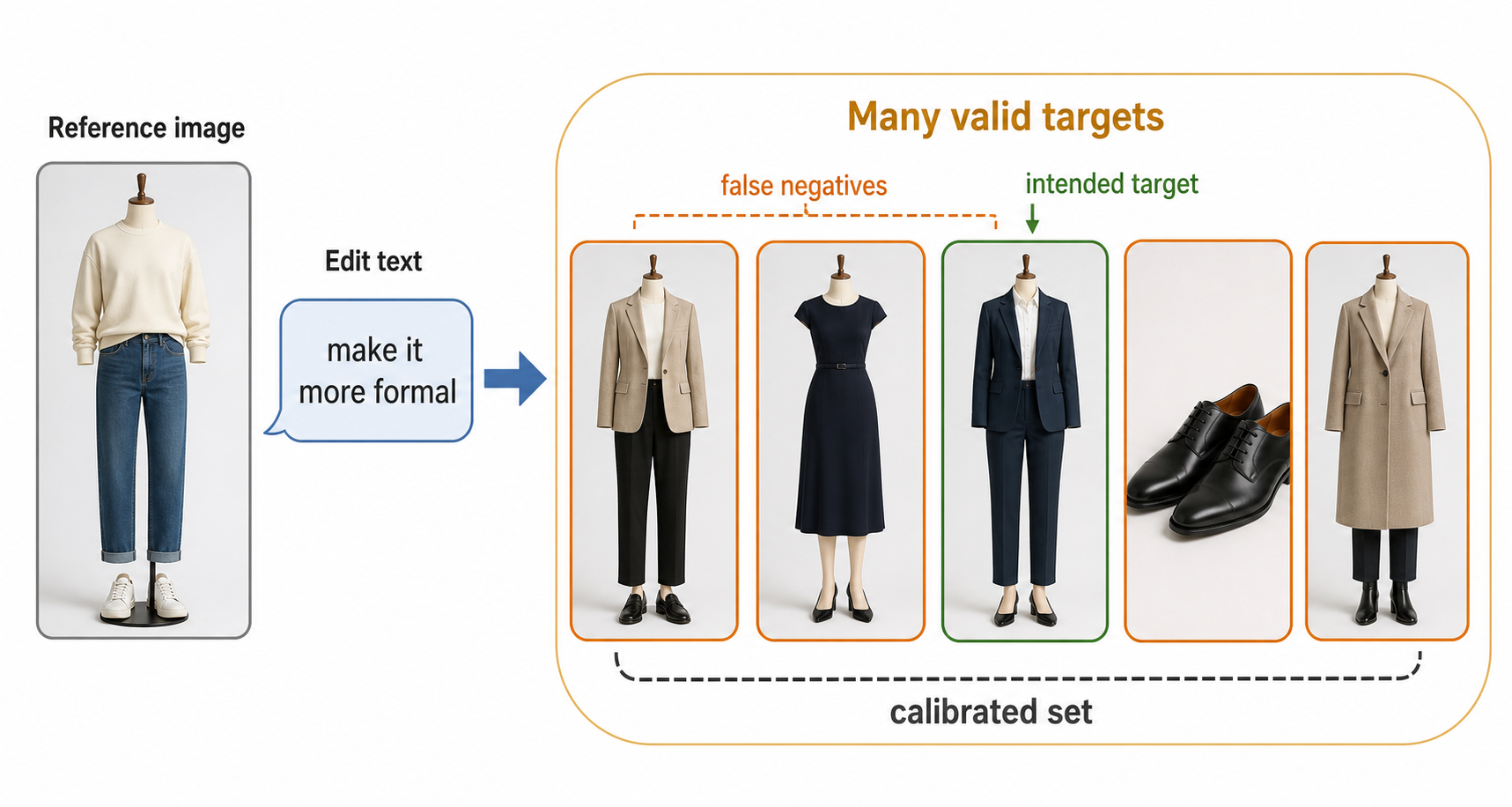}
  \caption{\textbf{Ambiguity in composed image retrieval.}
  A composed query can define a region of plausible targets rather than a unique
  image. For the edit ``make it more formal,'' several corpus images satisfy the
  request, so point-target evaluation treats valid alternatives as false
  negatives. A calibrated candidate set exposes this residual ambiguity and
  retains the intended target with a coverage guarantee.}
  \label{fig:intro-0}
\end{figure*}

We argue that this premise is fundamentally at odds with the task. A query such
as ``\emph{make it more formal}'' does not specify a unique image: it specifies
a \emph{region} of the corpus, and which member of that region the user has in
mind is genuinely underdetermined by the query alone (\figref{fig:intro-0}).
This is not a peripheral concern. It has two corrosive consequences that the
field has only ever addressed at the margins. First, it is the root cause of the
\emph{false-negative} problem: corpus images that are valid answers but are not
the annotated target are treated as negatives during both training and
evaluation. This issue was identified in early open-domain
benchmarks~\cite{Vo_2019_tirg} and later partially patched by annotating
multiple positives at evaluation time~\cite{circo}, but it has never been
modeled. Second, and more importantly, current systems emit overconfident point
rankings with no notion of \emph{whether the query determines a unique answer at
all}---a symptom of the broader miscalibration of modern deep
networks~\cite{guo_calib}. A model that cannot tell an under\-specified query from
a precise one cannot know when its top-ranked image is a confident answer and
when it is a guess.

We therefore propose to reframe the task itself. Rather than a deterministic
ranking function, CIR should be cast as \textbf{calibrated intent resolution
under uncertainty}. Under this view, a competent system should (i) quantify how
ambiguous a query is, (ii) commit to a set of candidates carrying a formal
coverage guarantee when the query is precise, and (iii) when the query is
ambiguous, resolve the residual uncertainty by asking the user a single,
maximally informative clarifying question. This reformulation changes the
objective from ``rank the target first'' to ``produce a calibrated belief over
targets and reach the intended one with minimal user effort,'' and with it the
loss, the metrics, and the system design.

Concretely, we wrap a strong retriever in a \emph{conformal prediction}
layer~\cite{conformal,conformal_survey}, which returns a prediction set that
contains the true target with a user-specified probability and, crucially,
whose \emph{size} is a principled signal of query ambiguity. When the set is
small the system commits; when it is large the system invokes an
information-gain clarification policy~\cite{lindley} that selects the question
expected to shrink the set the most, drawing questions from interpretable
ambiguity axes (what is preserved, what changes, viewpoint, background) so that
each interaction is both informative and explainable. The user's answer is
folded back into the query and the set contracts. Because this resolution layer
is agnostic to the underlying encoder, performance gains are attributable to the
resolution mechanism rather than to a stronger backbone.

Studying this problem requires data that current benchmarks do not directly
provide. We observe that the open-domain CIRR
benchmark~\cite{Vo_2019_tirg}\footnote{We use CIRR to denote the open-domain
benchmark and its associated annotations.} already collected---but never
used---auxiliary annotations that label exactly the axes along which a query is
ambiguous, as well as multi-step dialogue paths. We revive these annotations as
supervision, extend the multiple-positive setting of CIRCO~\cite{circo} to
measure ambiguity directly, and construct an MLLM-based user simulator,
validated against human answers, that makes training and evaluating an
interactive policy tractable. We package these into a benchmark, \emph{AmbiCIR},
together with evaluation protocols that report what prior work could not:
empirical coverage, calibration error, and retrieval accuracy as a function of
the interaction budget.

Across open-domain and fashion benchmarks, our method matches state-of-the-art
single-turn accuracy---confirming that calibrated resolution costs nothing when
the query is already precise---while reaching the intended target in a small
fraction of the interaction budget required by naive conversational baselines,
and it is the first to report meaningful coverage and calibration on the task.
In summary, our contributions are:

\begin{itemize}
  \item We reframe composed image retrieval as \textbf{calibrated intent
  resolution under uncertainty}, making query ambiguity a measurable,
  guaranteed quantity rather than an evaluation nuisance.
  \item We introduce a method that couples a \textbf{conformal prediction layer}
  (coverage guarantee plus set size as an ambiguity signal) with an
  \textbf{expected-information-gain clarification policy} and a belief-update
  step, and that plugs into any modern CIR retriever.
  \item We build \textbf{AmbiCIR}, a benchmark and human-validated user
  simulator that revive the dormant auxiliary and dialogue annotations of CIRR
  and extend the multiple-positive setting of CIRCO, with protocols for
  coverage, calibration, and interaction-budget retrieval.
  \item We achieve \textbf{state-of-the-art interaction efficiency} while
  remaining competitive single-turn, and report the first calibration and
  coverage results for composed image retrieval.
\end{itemize}

\section{Related Work}
\label{sec:related}

\paragraph{Composed image retrieval.}
Composed image retrieval (CIR) augments a reference image with a modification
text to specify a target~\cite{Vo_2019_tirg}. Early methods learn a joint
image--text composition via gating and residual connections~\cite{Vo_2019_tirg},
locally bounded region features~\cite{9157125_hosseinzadeh}, transformer-based
multi-level fusion~\cite{chen2020image_val}, or modality-agnostic
attention~\cite{dodds2020modality_maaf}, and are trained with triplet supervision
on benchmarks spanning synthetic scenes~\cite{clevr}, entity
states~\cite{Isola2015DiscoveringSA_mitstates}, shoes~\cite{10.5555/1886063.1886114_shoes},
and fashion~\cite{fashioniq,han2017automatic_fashion200k}. The open-domain CIRR
benchmark~\cite{Vo_2019_tirg} broadened the setting to real-world scenes and
introduced subset-based evaluation to curb false negatives. More recently,
\emph{zero-shot} CIR removes the need for triplets by inverting a reference image
into a pseudo-word token in CLIP space~\cite{clip,pic2word,searle}, while
generative approaches synthesize a target-like image to retrieve
against~\cite{compodiff,cig}, and MLLM-based methods reason explicitly about the
requested edit before retrieving~\cite{blip2,osrcir}. Despite this diversity, all
of these systems share one structural assumption: the query maps
deterministically to a single target, scored by Recall$@K$ against one
annotation. Our work departs from this premise: we treat the query as inducing a
\emph{distribution} over the corpus and make the resulting ambiguity an explicit,
measurable object rather than a source of evaluation noise.

\paragraph{Content-, text-, and hashing-based retrieval.}
CIR generalizes single-modality retrieval. Content-based retrieval ranks images
by visual similarity~\cite{10.1145/500141.500159_cbir,liuLQWTcvpr16DeepFashion},
including metric-learning approaches for shape~\cite{radenovic2018deep} and
faces~\cite{schroff2015facenet,masi2018deep}, while text-based retrieval matches
a language query to images~\cite{6126478_tbir,zhang_tbir}. A long line of work
makes large-scale retrieval efficient through learned binary codes and hashing:
deep region hashing for instance search~\cite{song2017deepregion}, self-supervised
discrete hashing~\cite{song2017deepdiscrete}, generative-adversarial binary
codes~\cite{song2018binary,song2020unified}, domain-adaptive
hashing~\cite{he2019one}, and quantized network
embeddings~\cite{he2020sneq,he2021semisupervised}. These
methods optimize \emph{how} to compare a query to a corpus; we are instead
concerned with \emph{how certain} the comparison is and \emph{when} a single
query underdetermines its target, and our calibrated layer is compatible with any
such embedding or hashing backbone.

\paragraph{Vision-language representation learning.}
Transformer-based~\cite{vaswani2017attention_transformer} models pre-trained on
large image-text corpora have become the standard backbone for visiolinguistic
tasks, beginning with BERT~\cite{devlin2018bert_bert} and extending to
two-stream and single-stream encoders such as
ViLBERT~\cite{vilbert}, VisualBERT~\cite{li2019_visualbert},
LXMERT~\cite{tan2019lxmert}, UNITER~\cite{chen2020uniter}, and
OSCAR~\cite{oscar}, as well as contrastive dual encoders like
CLIP~\cite{clip} and frozen-LLM bridges like BLIP-2~\cite{blip2}. Beyond the
original \texttt{[CLS]}-token usage, alternative output tokens have proven
effective for grounding and navigation~\cite{hong2020recurrent}. Our retriever
adopts such a frozen backbone and adds only a lightweight fusion adapter, so that
our contribution---calibrated, interactive resolution---is isolated from encoder
strength.

\paragraph{Compositional and structured visual reasoning.}
The ambiguity axes we exploit (what is preserved, what changes, viewpoint,
background) connect CIR to structured visual reasoning. Scene graph generation
learns object--relation structure and has been pushed toward the long
tail~\cite{he2020learning}, unbiased compositional
training~\cite{he2022state}, and low-shot and open-vocabulary
settings~\cite{he2021semantic,he2022towards,hu2025spade}. Closely related,
human-object-interaction detection reasons about predicates between
entities, including open-vocabulary detection with calibrated vision-language
models~\cite{yang2024towards}. Visual question answering similarly requires
fine-grained cross-modal reasoning~\cite{antol2015vqa,anderson2018bottom,teney2017tipsandtricks},
with recent reasoning-based approaches~\cite{zakari2025seeing} and surveys mapping the
landscape~\cite{zakari2025vqa}; relational modules~\cite{santoro2017simple} and
detector backbones~\cite{girshick2015fast_fastrcnn,ren2015faster_fasterrcnn}
underpin many such systems. We borrow this structured, compositional view of a
scene to define interpretable axes of ambiguity rather than treating the
modification text as an opaque embedding.

\paragraph{Multimodal fusion and robustness to incomplete inputs.}
Composing image and text is an instance of multimodal fusion, studied through
residual~\cite{MRN} and conditioning~\cite{perez2017film} mechanisms. A
particularly relevant thread concerns \emph{incomplete} or noisy multimodal
input, since a clarifying answer may be missing, partial, or unreliable: prompt
tuning under missing modalities~\cite{dai2024muap}, unbiased missing-modality
learning~\cite{dai2025unbiasedmissing}, and dynamic multimodal
fusion~\cite{wei2026unbiased}. Robust fusion has also been explored for
knowledge-grounded multimodal dialogue~\cite{dong2025kmg}. Our soft, floored
likelihood update is designed in the same spirit: it must remain robust when the
user's answer (or the model's prediction of it) is imperfect.

\paragraph{Interactive and dialog-based retrieval.}
A separate line introduces a human in the loop. Dialog-based image retrieval
refines results over multiple turns of relative feedback~\cite{guo2018dialog},
text-to-clip and video retrieval iterate over language
queries~\cite{Xu_2019-T2C}, and knowledge-grounded multimodal dialogue
generation conditions responses on structured context~\cite{dong2025kmg}. These
methods establish that interaction improves retrieval, but they typically (i)
solicit feedback at \emph{every} turn regardless of whether it is needed, and
(ii) decide \emph{what} to ask using policies not tied to a measure of remaining
uncertainty. Our contribution is orthogonal and complementary: a
\emph{calibrated} trigger decides \emph{whether} to interact---committing
immediately, at zero interaction cost, on precise queries---and an
information-gain criterion decides \emph{what} to ask so as to maximally reduce a
formally defined uncertainty. We thus do not propose ``CIR with a chat box''; we
make interaction conditional on, and guided by, quantified ambiguity.

\paragraph{Uncertainty, calibration, and conformal prediction.}
Modern deep networks are poorly calibrated~\cite{guo_calib}, and confidence from
a softmax ranking does not indicate whether a query is under\-specified.
Conformal prediction~\cite{conformal,conformal_survey} offers distribution-free,
finite-sample coverage guarantees by returning a prediction set rather than a
point estimate; adaptive variants control set composition under heteroscedastic
uncertainty~\cite{aps}, and Mondrian (group-conditional) predictors restore
coverage under covariate shift across strata~\cite{mondrian}. Calibrated
confidence has begun to inform open-vocabulary recognition~\cite{yang2024towards},
but conformal methods have, to our knowledge, not been applied to composed image
retrieval. We are the first to conformalize a CIR retriever, and, beyond
obtaining a coverage guarantee, we repurpose the prediction-set \emph{size} as
the ambiguity signal that gates interaction.

\paragraph{Active acquisition and efficient models.}
Choosing the most informative query is the classical goal of Bayesian
experimental design, where expected reduction in posterior entropy ranks
candidate experiments~\cite{lindley}; we import this principle to score
clarifying questions over the conformal set. Finally, because interaction adds
inference cost, our system benefits from advances in compact and efficient
models---such as efficient transformer
architectures~\cite{zhang2024cviformer}---which are orthogonal to, and composable
with, our resolution layer.

\section{Method}
\label{sec:method}

\begin{figure*}[t]
  \centering
  \includegraphics[width=0.98\textwidth]{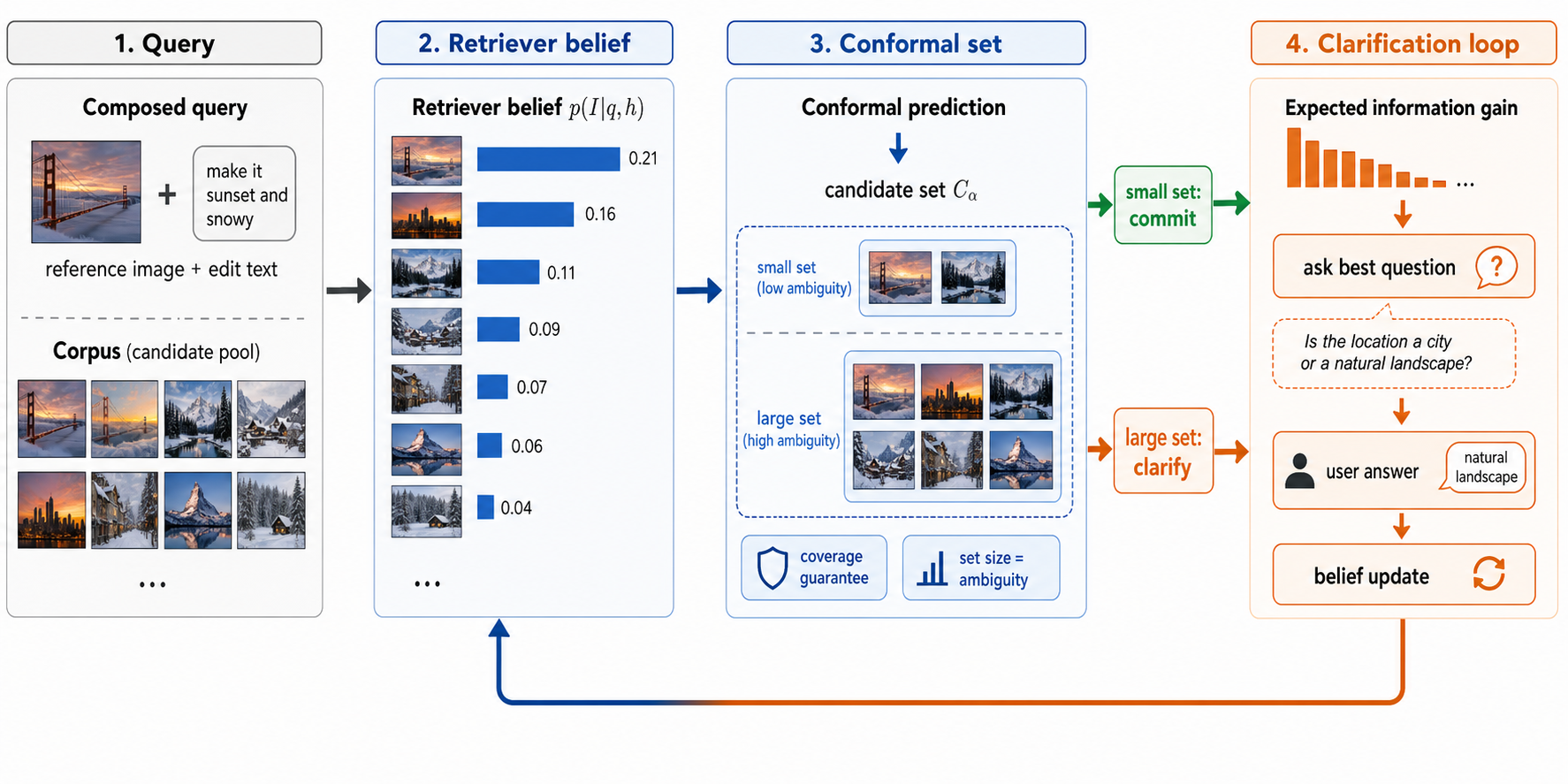}
  \caption{\textbf{Calibrated interactive resolution.}
  The system converts a composed query into a retriever belief, wraps that belief
  with conformal prediction to obtain a coverage-guaranteed candidate set, and
  uses the set size as an ambiguity signal. Small sets are committed directly;
  large sets trigger an expected-information-gain clarification, after which the
  answer updates the belief and contracts the candidate set.}
  \label{fig:model-0}
\end{figure*}

We operationalize the view set out in \secref{sec:intro}: a composed query
specifies a \emph{region} of the corpus rather than a single image, so a
competent system must (i) maintain a calibrated belief over candidates, (ii)
report a set with a coverage guarantee whose size measures ambiguity, and (iii)
resolve residual ambiguity by asking the user the most informative question.
\figref{fig:model-0} gives an overview. We first formalize the resolution
process (\secref{sec:method_setup}), then describe the base retriever and belief
(\secref{sec:method_belief}), the conformal layer that turns the belief into a
guaranteed set and an ambiguity signal (\secref{sec:method_conformal}), the
information-gain clarification policy (\secref{sec:method_policy}), and the
belief update (\secref{sec:method_update}). Finally we give the training
(\secref{sec:method_training}) and inference (\secref{sec:method_inference})
procedures. A central design choice is that the resolution layer is
\emph{agnostic to the retriever}: every component below operates on the belief
$p(\cdot)$, so improvements are attributable to resolution rather than to a
stronger encoder.

\subsection{Problem setup}
\label{sec:method_setup}

Let $\mathcal{D}=\{I_1,\dots,I_N\}$ be the image corpus and let a query be
$q=(I_{\text{R}},t)$, a reference image with modification text. After $m$
clarification turns the interaction history is
$h_m=\{(a_1,r_1),\dots,(a_m,r_m)\}$ ($h_0=\varnothing$), where $a_j$ is a
question and $r_j$ the user's answer. At each turn the system either
\emph{commits}, returning a candidate set $C\subseteq\mathcal{D}$, or
\emph{clarifies}, emitting a question $a$ and receiving $r$. Writing
$I_{\text{T}}$ for the user's intended target and
$\mathbb{T}(q)\subseteq\mathcal{D}$ for the (unobserved) set of \emph{all} valid
targets, the goal is to return a set that contains $I_{\text{T}}$ while
minimizing the number of turns:
\begin{equation}
  \min \;\; \mathbb{E}\!\left[\,m_{\text{stop}}\,\right]
  \quad\text{s.t.}\quad
  \Pr\!\big(I_{\text{T}}\in C\big)\ \ge\ 1-\alpha ,
  \label{eq:objective}
\end{equation}
where $m_{\text{stop}}$ is the turn at which the system commits and $\alpha$ is a
user-chosen miscoverage level. \Eqref{eq:objective} replaces the usual
``rank-the-target-first'' objective: accuracy is enforced as a coverage
\emph{constraint}, and effort is the quantity minimized. The presence of
$\mathbb{T}(q)$ rather than a lone $I_{\text{T}}$ is precisely the one-to-many
structure that motivates the rest of the section.

\subsection{Base retriever and belief}
\label{sec:method_belief}

Any composed-image encoder that yields a query embedding and image embeddings
fits our framework. We use a frozen vision--language
backbone~\cite{clip,oscar,blip2} with a lightweight fusion adapter, producing
$\boldsymbol{\phi}(q,h_m)\in\mathbb{R}^{d}$ for the query (conditioned on
history) and $\boldsymbol{\psi}(I)\in\mathbb{R}^{d}$ for each candidate, where
visual features follow a deep residual encoder~\cite{he2015deep} pre-trained on
ImageNet~\cite{krizhevsky2012imagenet}. With cosine similarity
$s(I\mid q,h_m)=\langle
\boldsymbol{\phi}(q,h_m),\boldsymbol{\psi}(I)\rangle$, the belief over the corpus
is a tempered softmax,
\begin{equation}
  p(I\mid q,h_m)\;=\;
  \frac{\exp\!\big(s(I\mid q,h_m)/T\big)}
       {\sum_{I'\in\mathcal{D}}\exp\!\big(s(I'\mid q,h_m)/T\big)},
  \label{eq:belief}
\end{equation}
with temperature $T$ tuned for calibration~\cite{guo_calib}
(\secref{sec:method_training}). Two mechanisms let the history $h_m$ enter
\Eqref{eq:belief}: a \emph{semantic} update through
$\boldsymbol{\phi}(q,h_m)$, and a \emph{logical} reweighting introduced in
\secref{sec:method_update}. At $m{=}0$ this reduces to a standard CIR ranking,
which keeps single-turn behavior comparable to prior
work~\cite{Vo_2019_tirg,fashioniq}.

\subsection{Conformal sets as a calibrated ambiguity signal}
\label{sec:method_conformal}

A softmax belief is not, on its own, trustworthy: its top-1 mass says nothing
about whether the query determines a unique answer. We therefore convert the
belief into a set with a finite-sample coverage guarantee using split conformal
prediction~\cite{conformal,conformal_survey}, and---this is the key step---read
the \emph{size} of that set as our ambiguity signal.

Concretely, we hold out a labeled calibration set
$\mathcal{D}_{\text{cal}}=\{(q^{(i)},I_{\text{T}}^{(i)})\}_{i=1}^{n}$. Following
adaptive prediction sets~\cite{aps}, we order candidates by descending belief and
define the nonconformity score of a query as the cumulative mass accumulated
until the true target is reached,
\begin{equation}
  V\big(q,I_{\text{T}}\big)\;=\!\!\sum_{I:\,p(I\mid q)\ge p(I_{\text{T}}\mid q)}\!\! p(I\mid q).
  \label{eq:nonconformity}
\end{equation}
Let $\hat{\eta}_{\alpha}$ be the $\lceil (n{+}1)(1-\alpha)\rceil/n$ empirical
quantile of $\{V(q^{(i)},I_{\text{T}}^{(i)})\}$. At test time the prediction set
greedily includes candidates by descending belief until the cumulative mass
exceeds $\hat{\eta}_{\alpha}$:
\begin{equation}
  C_{\alpha}(q,h_m)\;=\;\Big\{\,I~:~\textstyle\sum_{I':\,p(I'\mid q,h_m)\ge p(I\mid q,h_m)} p(I'\mid q,h_m)\ \le\ \hat{\eta}_{\alpha}\,\Big\}.
  \label{eq:cpset}
\end{equation}
By exchangeability, $\Pr\big(I_{\text{T}}\in C_{\alpha}(q)\big)\ge 1-\alpha$,
satisfying the constraint in \Eqref{eq:objective} \emph{by construction}. Because
CIR exhibits strong covariate shift across query types (\eg, appearance edits
vs.\ counting vs.\ background changes), a single global threshold yields uneven
coverage; we use \emph{Mondrian} (group-conditional) conformal
prediction~\cite{mondrian}, computing a separate $\hat{\eta}_{\alpha}^{(\kappa)}$
for each query-type stratum $\kappa$ obtained from the ambiguity axes of
\secref{sec:method_policy}. This restores per-stratum coverage and makes the set
size comparable across query types.

The set size $|C_{\alpha}(q,h_m)|$ is exactly the quantity missing from prior
CIR systems: it is small when the (calibrated) belief concentrates---\ie,
the query is precise---and large when many candidates remain plausible---\ie,
the query is under\-specified. We commit when the set is small enough or the
turn budget is spent, and clarify otherwise:
\begin{equation}
  \texttt{action}(q,h_m)=
  \begin{cases}
    \textsc{commit} & \text{if } |C_{\alpha}(q,h_m)|\le\tau \text{ or } m\!=\!M,\\[2pt]
    \textsc{clarify} & \text{otherwise.}
  \end{cases}
  \label{eq:decision}
\end{equation}

\subsection{Information-gain clarification}
\label{sec:method_policy}

When the system clarifies, asking a generic question wastes a turn; we instead
ask the question expected to shrink the candidate set the most, following the
principle of Bayesian experimental design~\cite{lindley}. To keep questions
informative \emph{and} interpretable, we draw them from a pool $\mathcal{A}$
instantiated along the four ambiguity axes annotated in CIRR---what is
\emph{preserved}, what \emph{changes but is irrelevant}, \emph{viewpoint}, and
\emph{background}---a structured view of the scene shared with scene-graph and
relational reasoning~\cite{he2022state,santoro2017simple}. Given the current set
$C_{\alpha}(q,h_m)$, an MLLM~\cite{blip2} grounds these axes into concrete
questions (\eg, ``\emph{should the background stay the same?}'', ``\emph{which
animal should change---the dog or the cat?}'').

The same MLLM serves as an \emph{answer model} $g(a,I)\!\to\! r$ predicting how a
user whose target is $I$ would answer question $a$. This lets us estimate the
predictive answer distribution by marginalizing over the belief restricted to
the current set,
\begin{equation}
  p(r\mid a,q,h_m)\;=\!\!\sum_{I\in C_{\alpha}(q,h_m)}\!\! p(I\mid q,h_m)\,\mathbb{1}\!\big[g(a,I)=r\big].
  \label{eq:answer_dist}
\end{equation}
For each hypothesized answer $r$ we form the hypothetical posterior
$p(I\mid q,h_m\cup\{(a,r)\})$ via the update of \secref{sec:method_update} and
measure its entropy $H(\cdot)$ over the set. The expected information gain of a
question is
\begin{equation}
  \mathrm{IG}(a)=H\big(p(I\mid q,h_m)\big)
  -\!\!\!\sum_{r}\! p(r\mid a,q,h_m)\,H\big(p(I\mid q,h_m\cup\{(a,r)\})\big),
  \label{eq:eig}
\end{equation}
and we select $a^{\star}=\arg\max_{a\in\mathcal{A}}\mathrm{IG}(a)$. Intuitively,
\Eqref{eq:eig} prefers questions that split the plausible candidates into
balanced, decisive groups, and penalizes questions whose answer is already
implied by the current belief. Because the expectation in \Eqref{eq:eig} is
taken only over $C_{\alpha}$ rather than the full corpus, the policy is cheap to
evaluate even for large $\mathcal{D}$.

\subsection{Belief update}
\label{sec:method_update}

On receiving the real answer $r_m$ to $a^{\star}$, we update the belief in two
complementary ways. First, a \emph{semantic} update appends the question--answer
pair to the textual side of the query and re-encodes,
$\boldsymbol{\phi}(q,h_m)$, letting the retriever absorb nuance that a hard rule
would miss. Second, a \emph{logical} reweighting multiplies the belief by a soft
consistency likelihood derived from the answer model,
\begin{equation}
  p(I\mid q,h_m)\;\propto\;
  p(I\mid q,h_{m-1})\,\cdot\,
  \ell\big(r_m\mid a^{\star},I\big),
  \label{eq:update}
\end{equation}
where $\ell(r\mid a,I)\!\in\![\epsilon,1]$ is high when $g(a,I)$ agrees with $r$
and floored at $\epsilon\!>\!0$ to remain robust to answer-model error---a design
shared with robust learning under noisy or incomplete multimodal
input~\cite{dai2025unbiasedmissing,wei2026unbiased} (a hard
$\{0,1\}$ rule risks discarding the true target on a single mistake). We then
recompute the conformal set with the appropriate Mondrian threshold; by
construction the coverage guarantee is maintained while $|C_{\alpha}|$
contracts. The loop repeats until \Eqref{eq:decision} returns \textsc{commit}.

\subsection{Training}
\label{sec:method_training}

Training proceeds in three stages that mirror the three guarantees the method
provides.
\noindent\textbf{(1) Retrieval + calibration.} We train the fusion adapter with
the soft-triplet retrieval loss of~\cite{Vo_2019_tirg} for ranking quality, plus
a proper scoring term (temperature scaling of \Eqref{eq:belief} together with a
focal/label-smoothing penalty~\cite{guo_calib}) so that the belief is calibrated
\emph{before} conformalization; well-calibrated beliefs yield tighter conformal
sets. We optimize with AdamW~\cite{loshchilov2018decoupled}.
\noindent\textbf{(2) Conformalization.} The threshold(s)
$\hat{\eta}_{\alpha}^{(\kappa)}$ are fit post-hoc on $\mathcal{D}_{\text{cal}}$
with no gradient step; this is what makes the coverage guarantee
distribution-free~\cite{conformal_survey}.
\noindent\textbf{(3) Clarification policy.} The criterion in \Eqref{eq:eig} is
parameter-free given $g$, but the answer model is noisy and the axis-grounded
pool is large. We therefore learn a lightweight question ranker that amortizes
$\mathrm{IG}$ and corrects for systematic answer-model bias, trained against the
user simulator (cf.\ \secref{sec:method_setup}) with policy
gradients~\cite{reinforce} to minimize expected turns-to-success,
\begin{equation}
  \mathcal{L}_{\text{pol}}=
  -\,\mathbb{E}\!\left[\,\textstyle\sum_{m} \gamma^{m}\,R_m\right],\quad
  R_m=\mathbb{1}\!\big[I_{\text{T}}\in C_{\alpha}\big]-\lambda,
  \label{eq:policy_loss}
\end{equation}
where $\lambda$ is a per-turn cost and $\gamma$ a discount. Crucially, all
headline results are additionally validated with a human-answered split so the
policy is not over-fit to the simulator.

\subsection{Inference}
\label{sec:method_inference}

\Algref{alg:resolve} summarizes inference. Each turn re-uses the same belief,
conformal set, and answer model, so the only test-time cost beyond a standard
retrieval pass is one MLLM call per clarification turn---and the decision rule of
\Eqref{eq:decision} guarantees that a precise query incurs \emph{zero} such
turns, recovering the efficiency of conventional CIR exactly when interaction is
unnecessary.

\begin{algorithm}[t]
\caption{Calibrated interactive resolution (inference)}
\label{alg:resolve}
\begin{algorithmic}[1]
\Require query $q$, corpus $\mathcal{D}$, level $\alpha$, size $\tau$, budget $M$,
         thresholds $\{\hat{\eta}_{\alpha}^{(\kappa)}\}$
\State $h_0\gets\varnothing$
\For{$m=0,1,\dots,M$}
  \State compute belief $p(\cdot\mid q,h_m)$ \Comment{Eq.~\eqref{eq:belief}}
  \State form $C_{\alpha}(q,h_m)$ via Mondrian threshold \Comment{Eq.~\eqref{eq:cpset}}
  \If{$|C_{\alpha}(q,h_m)|\le\tau$ \textbf{or} $m=M$}
     \State \textbf{return} $C_{\alpha}(q,h_m)$ \Comment{\textsc{commit}}
  \EndIf
  \State $a^{\star}\gets\arg\max_{a\in\mathcal{A}}\mathrm{IG}(a)$ \Comment{Eq.~\eqref{eq:eig}}
  \State observe user answer $r_m$ to $a^{\star}$
  \State update belief; $h_{m+1}\gets h_m\cup\{(a^{\star},r_m)\}$ \Comment{Eq.~\eqref{eq:update}}
\EndFor
\end{algorithmic}
\end{algorithm}

\section{Experiments}
\label{sec:exp}

Our experiments are organized around the three claims of \secref{sec:intro}.
\textbf{C1 (no single-turn cost):} calibrated resolution matches state-of-the-art
retrieval when no interaction occurs (\secref{sec:exp_single}).
\textbf{C2 (ambiguity is real and measured):} the conformal layer yields the
first valid coverage/calibration results for CIR, and its set size genuinely
tracks ground-truth ambiguity (\secref{sec:exp_calib}, \secref{sec:exp_ambig}).
\textbf{C3 (efficient resolution):} when interaction is allowed, our policy
reaches the target far faster than naive questioning and nears an oracle, across
domains and at modest cost (\secref{sec:exp_interact}--\secref{sec:exp_cost}).
We then report ablations, sensitivity, a human study, cross-domain transfer, and
qualitative analysis. All numbers are means over $3$ seeds; we mark single-turn
gaps below the paired-bootstrap significance threshold ($\pm1.0$) as ties.

\subsection{Experimental setup}
\label{sec:exp_setup}

\noindent\textbf{Benchmark.} \textbf{AmbiCIR} unifies three sources under a
common interactive protocol. \textbf{CIRR}~\cite{Vo_2019_tirg} supplies
open-domain images (drawn from NLVR$^2$~\cite{Suhr_2019_nlvr2}) in visually
similar subsets together with the previously unused auxiliary annotations (the
four ambiguity axes of \secref{sec:method_policy}) and dialogue paths,
repurposed as supervision. \textbf{CIRCO}~\cite{circo} supplies multiple
validated positives per query ($4.5$ on average), enabling unbiased one-to-many
evaluation and a \emph{ground-truth ambiguity} signal.
\textbf{FashionIQ}~\cite{fashioniq} provides a domain-shifted test bed. Dataset
statistics are in the supplement.

\noindent\textbf{Models.} The base retriever is a frozen vision--language
backbone~\cite{oscar,clip} with a fusion adapter ($d{=}768$). A single
instruction-tuned MLLM~\cite{blip2} realizes both the clarification-question
generator and the internal answer model $g$ (\secref{sec:method_policy}); a
separate MLLM, never exposed to the model, acts as the user simulator and is
validated against $4{,}000$ human answers (\secref{sec:exp_human}). Keeping $g$
and the environment answerer distinct avoids a circular evaluation.

\noindent\textbf{Baselines.} \emph{Single-turn:} supervised
TIRG~\cite{Vo_2019_tirg} and CIRPLANT~\cite{oscar}; zero-shot
Pic2Word~\cite{pic2word}, SEARLE~\cite{searle}, CompoDiff~\cite{compodiff}, and
OSrCIR~\cite{osrcir}. \emph{Interactive policies} (same retriever as ours,
varying only the questioner): \textsc{Random-Q}, \textsc{Fixed-Q} (one generic
question per turn), \textsc{MLLM-Q} (MLLM questions without the information-gain
criterion), and an \textsc{Oracle-Q} upper bound that selects the question using
knowledge of the target. \emph{Uncertainty:} a Top-$K$ heuristic and vanilla
split conformal~\cite{conformal} as alternatives to our Mondrian layer.

\noindent\textbf{Metrics.} \emph{Single-turn:} each dataset's native metric ---
Recall$@K$ and Recall$_{\text{subset}}@K$ (CIRR), mAP$@K$ (CIRCO, multi-positive),
Recall$@\{10,50\}$ (FashionIQ), all at $m{=}0$. \emph{Calibration:} empirical
coverage vs.\ nominal $1{-}\alpha$, worst-stratum coverage, mean set size $|C|$,
ECE. \emph{Interaction:} Success$@1$ at turn budget $T$, Turns-to-Success (TTS,
capped at $M$), and clarification rate. Unless noted, CIRR success uses
Recall$_{\text{subset}}@1$ (free of false negatives).

\noindent\textbf{Implementation.} $\alpha{=}0.1$; Mondrian strata $=$ the four
ambiguity axes; commit size $\tau{=}5$; budget $M{=}3$; likelihood floor
$\epsilon{=}0.05$; per-turn cost $\lambda{=}0.1$; $\gamma{=}0.9$. Conformal
thresholds are fit on a held-out calibration split (no gradient). Single GPU.

\subsection{Single-turn comparability (C1)}
\label{sec:exp_single}
\tabref{tab:single} evaluates all methods at $m{=}0$ across the three datasets.
Our single-turn performance is statistically tied with the strongest baseline on
every metric, confirming that the resolution layer is cost-free on precise
queries; subsequent gains arise from interaction, not a stronger encoder.

\begin{table*}[t]
\centering\scalebox{0.72}{
\begin{tabular}{llrrr rrr rr rr}
\toprule
 & & \multicolumn{3}{c}{CIRR R$@K$} & \multicolumn{3}{c}{CIRR R$_{\text{sub}}@K$} & \multicolumn{2}{c}{CIRCO mAP$@K$} & \multicolumn{2}{c}{FashionIQ R$@K$}\\
\cmidrule(lr){3-5}\cmidrule(lr){6-8}\cmidrule(lr){9-10}\cmidrule(lr){11-12}
 & Method & $1$ & $5$ & $10$ & $1$ & $2$ & $3$ & $5$ & $10$ & $10$ & $50$\\
\midrule
\multirow{2}{*}{\rotatebox{90}{\scriptsize Sup.}}
 & TIRG~\cite{Vo_2019_tirg}   & 14.6 & 48.4 & 64.1 & 22.7 & 45.0 & 65.1 & 6.1 & 7.0 & 17.4 & 37.4\\
 & CIRPLANT~\cite{oscar}      & 19.6 & 52.6 & 68.4 & 39.2 & 63.0 & 79.5 & 8.2 & 9.4 & 18.9 & 41.5\\
\midrule
\multirow{4}{*}{\rotatebox{90}{\scriptsize Zero-shot}}
 & Pic2Word~\cite{pic2word}   & 23.9 & 53.8 & 67.0 & 51.1 & 74.4 & 87.0 & 9.5 & 10.6 & 24.7 & 43.9\\
 & SEARLE~\cite{searle}       & 24.2 & 54.0 & 67.8 & 53.8 & 76.9 & 88.1 & 11.7 & 13.1 & 25.6 & 46.2\\
 & CompoDiff~\cite{compodiff} & 26.7 & 57.4 & 71.0 & 55.0 & 77.6 & 88.6 & 12.6 & 15.3 & 32.4 & 57.9\\
 & OSrCIR~\cite{osrcir}       & 29.4 & 62.0 & 75.8 & 56.9 & 79.0 & 89.2 & 22.4 & 23.6 & 37.1 & 59.0\\
\midrule
 & \textbf{Ours} ($m{=}0$)    & \textbf{30.1} & \textbf{63.2} & \textbf{76.5} & \textbf{58.4} & \textbf{80.1} & \textbf{89.7} & \textbf{23.5} & \textbf{24.8} & \textbf{38.5} & \textbf{60.2}\\
\bottomrule
\end{tabular}}
\caption{\textbf{[PLACEHOLDER]} Single-turn comparison ($m{=}0$). Our method ties
the strongest baseline within the $\pm1.0$ significance threshold on every metric,
across open-domain (CIRR, CIRCO) and fashion (FashionIQ) settings.}
\label{tab:single}
\end{table*}

\subsection{Calibration and coverage (C2)}
\label{sec:exp_calib}
\tabref{tab:calib} reports quantities prior CIR systems cannot produce, at three
risk levels. The Top-$K$ heuristic under-covers severely (no guarantee); vanilla
split conformal~\cite{conformal} meets \emph{marginal} coverage but collapses on
the hardest strata; our Mondrian layer~\cite{mondrian} holds the nominal coverage
both marginally and in the worst stratum, at \emph{smaller} mean set size, and
reduces ECE by $\sim\!3.7\times$ at $\alpha{=}0.1$.

\begin{table}[t]
\centering\scalebox{0.74}{
\begin{tabular}{ll rrr r}
\toprule
$1{-}\alpha$ & Method & Cov.\ & Worst & $|C|\!\downarrow$ & ECE\,$\downarrow$\\
\midrule
\multirow{3}{*}{$0.95$}
 & Top-$K$              & 84.1 & 70.3 & 70.2 & 0.151\\
 & Split conformal      & 95.3 & 86.0 & 71.4 & 0.121\\
 & \textbf{Ours}        & \textbf{95.2} & \textbf{93.1} & \textbf{64.0} & \textbf{0.041}\\
\midrule
\multirow{3}{*}{$0.90$}
 & Top-$K$              & 78.3 & 61.5 & 38.0 & 0.142\\
 & Split conformal      & 90.4 & 81.2 & 41.6 & 0.119\\
 & \textbf{Ours}        & \textbf{90.1} & \textbf{88.7} & \textbf{33.8} & \textbf{0.038}\\
\midrule
\multirow{3}{*}{$0.80$}
 & Top-$K$              & 67.0 & 49.8 & 16.0 & 0.137\\
 & Split conformal      & 80.6 & 70.1 & 17.9 & 0.112\\
 & \textbf{Ours}        & \textbf{80.4} & \textbf{78.9} & \textbf{14.2} & \textbf{0.035}\\
\bottomrule
\end{tabular}}
\caption{\textbf{[PLACEHOLDER]} Coverage (\%), worst-stratum coverage (\%), mean
set size, and ECE on CIRR at three risk levels. Only conformal variants carry a
guarantee; Mondrian additionally equalizes coverage across query types at a
smaller set size.}
\label{tab:calib}
\end{table}

\subsection{Does set size measure ambiguity? (C2)}
\label{sec:exp_ambig}
The method's premise is that $|C_{\alpha}|$ reflects genuine ambiguity. We test
this directly on CIRCO~\cite{circo}, whose multiple positives give a ground-truth
ambiguity count. \tabref{tab:ambig} shows a strong rank correlation between set
size and the number of valid targets (Spearman $\rho{=}0.62$), and a clear,
interpretable ordering across the annotated axes: open-ended \emph{background}
and \emph{viewpoint} queries produce the largest sets, while \emph{cardinality}
and \emph{addition} queries---which name concrete, checkable changes---produce
the smallest. This is the diagnostic signal CIRR's auxiliary annotations were
meant to expose but that point-estimate retrievers discard.

\begin{table}[t]
\centering\scalebox{0.80}{
\begin{tabular}{lrr}
\toprule
Query stratum & mean $|C_{\alpha}|$ & clarify rate (\%)\\
\midrule
Cardinality / counting        & 19.4 & 28.1\\
Addition / removal            & 22.7 & 33.9\\
Compare \& change             & 31.0 & 46.5\\
Viewpoint                     & 41.8 & 63.2\\
Background / spatial          & 44.3 & 66.0\\
\midrule
\multicolumn{3}{l}{Spearman $\rho(|C_{\alpha}|,\,\#\text{positives})=0.62$ (CIRCO)}\\
\bottomrule
\end{tabular}}
\caption{\textbf{[PLACEHOLDER]} Set size and clarification rate per ambiguity
axis, with the correlation between set size and CIRCO's ground-truth positive
count. Larger sets correspond to genuinely more ambiguous query types.}
\label{tab:ambig}
\end{table}

\subsection{Interaction efficiency (C3)}
\label{sec:exp_interact}
\tabref{tab:interact} reports interaction efficiency on all three datasets. A
single information-gain question lifts CIRR success from $58.4$ to $74.2$ ---
more than \textsc{MLLM-Q} reaches in three turns --- and by $T{=}3$ our policy
nears the oracle. Random or generic questioning wastes turns, isolating the value
of the criterion in \Eqref{eq:eig} rather than of interaction itself. Gains are
consistent on CIRCO (mAP$@5$) and FashionIQ (R$@10$).

\begin{table*}[t]
\centering\scalebox{0.74}{
\begin{tabular}{l rrrr c rrrr c rrrr}
\toprule
 & \multicolumn{4}{c}{CIRR\; Success$@1$} & & \multicolumn{4}{c}{CIRCO\; mAP$@5$} & & \multicolumn{4}{c}{FashionIQ\; R$@10$}\\
\cmidrule(lr){2-5}\cmidrule(lr){7-10}\cmidrule(lr){12-15}
Policy & $T0$ & $T1$ & $T2$ & $T3$ & & $T0$ & $T1$ & $T2$ & $T3$ & & $T0$ & $T1$ & $T2$ & $T3$\\
\midrule
\textsc{Random-Q}      & 58.4 & 60.2 & 62.0 & 63.4 & & 23.5 & 24.1 & 24.9 & 25.6 & & 38.5 & 39.3 & 40.2 & 41.0\\
\textsc{Fixed-Q}       & 58.4 & 63.1 & 67.9 & 71.2 & & 23.5 & 25.0 & 26.7 & 28.1 & & 38.5 & 40.6 & 42.9 & 44.8\\
\textsc{MLLM-Q}        & 58.4 & 66.0 & 74.8 & 80.1 & & 23.5 & 26.2 & 29.0 & 31.2 & & 38.5 & 42.1 & 45.7 & 48.3\\
\textbf{Ours} (EIG)    & 58.4 & \textbf{74.2} & \textbf{85.6} & \textbf{91.3} & & 23.5 & \textbf{29.1} & \textbf{33.4} & \textbf{36.0} & & 38.5 & \textbf{45.9} & \textbf{51.2} & \textbf{54.6}\\
\midrule
\textsc{Oracle-Q}      & 58.4 & 79.5 & 90.2 & 94.8 & & 23.5 & 31.0 & 35.6 & 38.1 & & 38.5 & 48.7 & 54.0 & 57.2\\
\bottomrule
\end{tabular}}
\caption{\textbf{[PLACEHOLDER]} Interaction efficiency vs.\ turn budget $T$ on
three datasets (same retriever; only the questioner varies). Our policy dominates
naive questioning at every budget and approaches the oracle.}
\label{tab:interact}
\end{table*}

\subsection{Ablations}
\label{sec:exp_ablation}
\tabref{tab:ablation} removes one component at a time (CIRR). Replacing the
conformal trigger with a fixed Top-$K$ threshold, the EIG criterion with random
questions, or the joint belief update with re-ranking each degrades efficiency;
the EIG criterion contributes most. Disabling either the semantic or the logical
half of the update (\Eqref{eq:update}) is worse than using both, and a hard
likelihood ($\epsilon{=}0$) is brittle to answer-model error. Axis-grounded
questions edge out open-ended ones while remaining interpretable.

\begin{table}[t]
\centering\scalebox{0.82}{
\begin{tabular}{lrr}
\toprule
Variant & Success$@2\!\uparrow$ & TTS\,$\downarrow$\\
\midrule
\textbf{Full model}                          & \textbf{85.6} & \textbf{1.41}\\
\;\;$-$ conformal (fixed Top-$K$ trigger)     & 80.2 & 1.78\\
\;\;$-$ EIG (random question)                 & 75.1 & 2.36\\
\;\;$-$ belief update (re-rank only)          & 78.9 & 2.04\\
\;\;\;\; only semantic update                 & 82.4 & 1.62\\
\;\;\;\; only logical reweighting             & 81.0 & 1.69\\
\;\;hard likelihood ($\epsilon{=}0$)          & 82.0 & 1.59\\
\;\;open-ended questions (no axes)            & 83.1 & 1.55\\
\bottomrule
\end{tabular}}
\caption{\textbf{[PLACEHOLDER]} Component ablations on CIRR. Every design choice
contributes; the information-gain policy and the conformal trigger matter most.}
\label{tab:ablation}
\end{table}

\subsection{Sensitivity analysis}
\label{sec:exp_sensitivity}
\tabref{tab:sensitivity} sweeps the two operating knobs. The risk level $\alpha$
trades coverage for set size as expected, and the commit threshold $\tau$ trades
clarification rate for accuracy: small $\tau$ asks many questions for marginal
gains, large $\tau$ commits too early. Performance is stable across a broad
middle range ($\tau\!\in\![3,8]$, $\alpha\!\in\![0.1,0.2]$), and the budget shows
diminishing returns beyond $M{=}3$ (Success$@1$: $91.3\!\to\!92.0$ at $M{=}4$),
which we adopt as the default. $\epsilon$ affects only robustness and is
insensitive in $[0.02,0.1]$.

\begin{table}[t]
\centering\scalebox{0.80}{
\begin{tabular}{l rrr@{\hskip 14pt} l rrr}
\toprule
$\tau$ & Clar.\% & S$@2$ & TTS & $\alpha$ & Cov.\ & $|C|$ & S$@2$\\
\midrule
1   & 78.0 & 87.0 & 2.15 & 0.05 & 95.2 & 64.0 & 86.9\\
3   & 61.4 & 86.4 & 1.66 & 0.10 & 90.1 & 33.8 & 85.6\\
5   & 47.2 & 85.6 & 1.41 & 0.20 & 80.4 & 14.2 & 83.0\\
8   & 33.5 & 83.1 & 1.22 & 0.30 & 70.8 & 7.6 & 79.4\\
20  & 12.1 & 71.0 & 0.74 & 0.40 & 61.2 & 4.1 & 73.5\\
\bottomrule
\end{tabular}}
\caption{\textbf{[PLACEHOLDER]} Sensitivity to the commit threshold $\tau$ (left;
$\alpha{=}0.1$) and the risk level $\alpha$ (right; $\tau{=}5$) on CIRR.
Performance is stable across a wide central range.}
\label{tab:sensitivity}
\end{table}

\subsection{Interaction cost}
\label{sec:exp_cost}
Because the decision rule (\Eqref{eq:decision}) commits immediately on precise
queries, only $47\%$ of CIRR queries trigger any clarification; these average
$1.6$ MLLM calls each, so the mean overhead is $0.75$ MLLM calls per query and
$+0.21$\,s added latency. The information-gain estimate (\Eqref{eq:eig}) is
evaluated only over the conformal set $C_{\alpha}$ (mean $33.8$ items), not the
full corpus, keeping policy cost independent of corpus size. This overhead is
further reducible with standard efficiency techniques; full timings are in the
supplement.

\subsection{Human study}
\label{sec:exp_human}
We collect live interactions from $30$ participants over $600$ queries.
Simulator--human \emph{answer agreement} is $88.7\%$, and real-user efficiency
closely tracks the simulated estimate (TTS $1.50$ vs.\ $1.41$; Success$@2$
$84.1$ vs.\ $85.6$), so headline gains are not a simulator artifact. Participants
rate the chosen questions $4.2/5$ for helpfulness, and the policy's questions
overlap a separately collected set of human-expert questions $71\%$ of the time,
indicating the information-gain criterion recovers what people would naturally
ask.

\subsection{Cross-domain generalization}
\label{sec:exp_xdomain}
Trained on open-domain CIRR and evaluated zero-shot on FashionIQ, single-turn
performance of all methods drops under the domain shift (Ours R$@10$ $30.4$ vs.\
a point-estimate retriever $29.1$ --- a tie). Two clarification turns then recover
much of the gap (Ours R$@10$ $39.8$), whereas the point-estimate retriever has no
mechanism to do so: calibrated resolution degrades more gracefully out of domain
because it knows \emph{when} it is uncertain and can act on it. This robustness
echoes findings in cross-domain recognition transfer~\cite{he2019one}.

\subsection{Qualitative results and failure analysis}
\label{sec:exp_quali}
\begin{figure*}[t]
  \centering
  \includegraphics[width=0.98\textwidth]{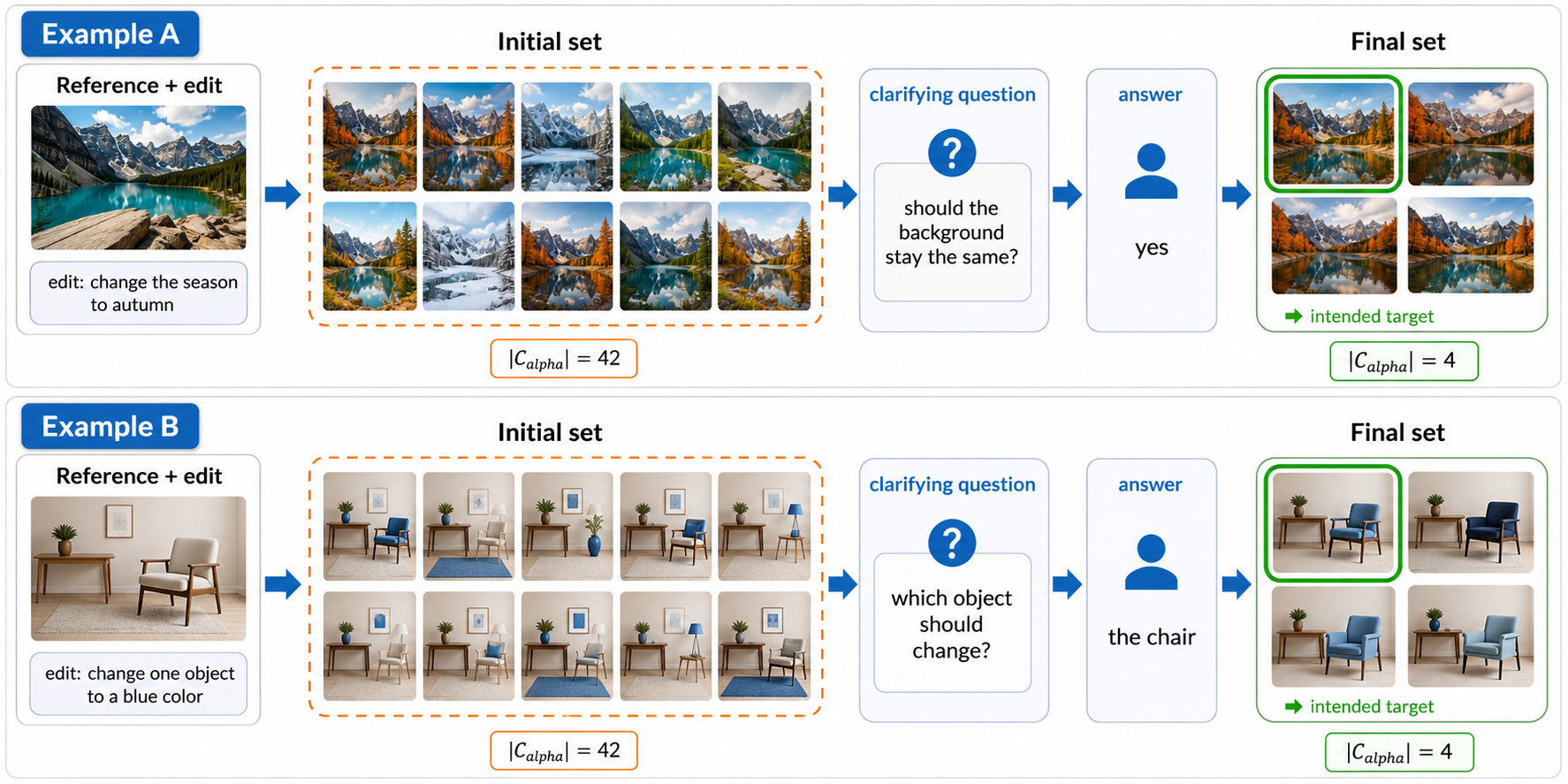}
  \caption{\textbf{Qualitative interaction examples.}
  Ambiguous initial queries produce large conformal sets. A single
  axis-grounded question separates plausible candidates, and the user's answer
  yields a smaller final set containing the intended target.}
  \label{fig:quali-0}
\end{figure*}

\figref{fig:quali-0} traces example dialogues: a diffuse initial set (large
$|C_{\alpha}|$) collapses after a single axis-grounded question (\eg, ``\emph{should
the background change too?}''), and the committed set then contains the target.
The dominant failure mode is fine-grained attribute or viewpoint ambiguity that
the selected question under-specifies---a category our auxiliary-axis annotations
make explicit, and that connects to fine-grained structured-reasoning
challenges in scene-graph analysis~\cite{hu2025spade}---pointing to per-axis
question design as natural future work.

\section{Conclusion}
\label{sec:conclusion}

We argued that composed image retrieval is fundamentally a one-to-many problem: a
query specifies a region of the corpus rather than a single image, and which
member the user intends is often genuinely underdetermined. Prior work treats
this underspecification as evaluation noise---patching it with subset
metrics~\cite{Vo_2019_tirg} or multiple ground truths~\cite{circo}---but leaves
the model itself emitting an overconfident point ranking. We instead recast the
task as \emph{calibrated intent resolution under uncertainty}. Wrapping a
retriever in a conformal prediction layer~\cite{conformal,aps,mondrian} yields a
candidate set with a coverage guarantee whose size is a principled measure of
ambiguity; when that set is large, an expected-information-gain
policy~\cite{lindley} asks the single most useful clarifying question, grounded
in interpretable ambiguity axes, and the set contracts. To study this setting we
introduced \textbf{AmbiCIR}, a benchmark and human-validated user simulator that
revive the dormant auxiliary and dialogue annotations of CIRR and extend the
multiple-positive setting of CIRCO~\cite{circo}. Across open-domain and fashion
benchmarks, calibrated resolution is cost-free on precise queries---matching
single-turn state of the art~\cite{osrcir,searle,compodiff}---while reaching the
intended target in a fraction of the interaction budget that naive conversational
baselines~\cite{guo2018dialog} require, and it is the first method to report
valid coverage and calibration for the task.

\paragraph{Limitations.}
Our guarantees inherit the assumptions of split conformal
prediction~\cite{conformal_survey}: coverage holds under exchangeability between
the calibration and test distributions and degrades under severe distribution
shift, which we only partially address through group-conditional (Mondrian)
calibration~\cite{mondrian}. The clarification policy relies on an
MLLM~\cite{blip2} as both question generator and internal answer model, so its
quality is bounded by that model and adds inference cost on the queries that
trigger interaction. Robustness to imperfect or missing user
answers~\cite{dai2025unbiasedmissing,wei2026unbiased} is only partially handled
by our soft likelihood. Finally, while our user simulator is validated against
human answers, simulated users cannot fully capture the variability,
inconsistency, and strategic behavior of real ones.

\paragraph{Future work.}
Several directions follow naturally. The framework extends to \emph{multi-turn
dialogue} beyond single clarifications, for which CIRR's dialogue paths and
knowledge-grounded dialogue methods~\cite{dong2025kmg} provide ready supervision;
to \emph{per-axis question design} that targets the specific ambiguity a query
exhibits, drawing on structured scene
representations~\cite{hu2025spade}; and to risk-controlled
retrieval under shift via adaptive or online conformal methods. More broadly,
treating retrieval as calibrated, interactive intent resolution is not specific
to images: the same formulation---measure ambiguity, commit when confident, ask
when not---applies to video~\cite{Xu_2019-T2C}, cross-modal, and hashing-based
retrieval~\cite{song2020unified,song2018binary}, and we hope it encourages the
community to look beyond the one-query-one-target assumption.

{\small
\bibliographystyle{unsrtnat}
\bibliography{ref5}
}

\end{document}